\documentclass[journal]{IEEEtran}
\usepackage{amsmath,amsfonts}
\usepackage{algorithmic}
\usepackage{algorithm}
\usepackage{array}
\usepackage[caption=false,font=normalsize,labelfont=sf,textfont=sf]{subfig}
\usepackage{textcomp}
\usepackage{stfloats}
\usepackage{url}
\usepackage{verbatim}
\usepackage{amsmath}
\usepackage{graphicx}
\usepackage{cite}
\usepackage{multirow}
\usepackage{booktabs} 
\usepackage{makecell} 
\usepackage{array}
\begin{document}

\title{Domain-invariant Representation Learning via Segment Anything Model for Blood Cell Classification}

\author{Yongcheng Li, Lingcong Cai, Ying Lu, Cheng Lin, Yupeng Zhang, Jingyan Jiang, Genan Dai, Bowen Zhang, Jingzhou Cao, Xiangzhong Zhang, and Xiaomao Fan
        % <-this % stops a space
\thanks{This work is partially supported by the Special subject on Agriculture and Social Development, Key Research and Development Plan in Guangzhou (2023B03J0172), the Basic and Applied Basic Research Project of Guangdong Province (2022B1515130009), and the Natural Science Foundation of Top Talent of SZTU (GDRC202318).}% <-this % stops a space
\thanks{Yongcheng Li, Lingcong Cai, Cheng Lin, Jingyan Jiang, Genan Dai, Bowen Zhang, Jingzhou Cao, and Xiaomao Fan are with the College of Big Data and Internet, Shenzhen Technology University, Shenzhen 518118, China (e-mail: vincentleen30@gmail.com; cailingcong@gmail.com; 2579858426@qq.com; astrofan2008@gmail.com). }
\thanks{Ying Lu and Xiangzhong Zhang are with the Department of Hematology, the Third Affiliated Hospital of Sun Yat-sen University,
Guangzhou, China (e-mail: luying5@mail.sysu.edu.cn; zhxzhong@mail.sysu.edu.cn).}
\thanks{Yupeng Zhang is with the School of Computer Science, South China Normal University, Guangzhou, China (zhangyupeng@m.scnu.edu.cn).}
\thanks{Yongcheng Li, Lingcong Cai, Cheng Lin, and Ying Lu contribute equally.}
\thanks{Xiaomao Fan is the corresponding author.}
}

\markboth{Journal of \LaTeX\ Class Files,~Vol.~14, No.~8, August~2021}%
{Shell \MakeLowercase{\textit{et al.}}: A Sample Article Using IEEEtran.cls for IEEE Journals}


\maketitle

\begin{abstract}
Accurate classification of blood cells is of vital significance in the diagnosis of hematological disorders. However, in real-world scenarios, domain shifts caused by the variability in laboratory procedures and settings, result in a rapid deterioration of the model's generalization performance. To address this issue, we propose a novel framework of domain-invariant representation learning (DoRL) via segment anything model (SAM) for blood cell classification. The DoRL comprises two main components: a LoRA-based SAM (LoRA-SAM) and a cross-domain autoencoder (CAE). The advantage of DoRL is that it can extract domain-invariant representations from various blood cell datasets in an unsupervised manner. Specifically, we first leverage the large-scale foundation model of SAM, fine-tuned with LoRA, to learn general image embeddings and segment blood cells. Additionally, we introduce CAE to learn domain-invariant representations across different-domain datasets while mitigating images' artifacts. To validate the effectiveness of domain-invariant representations, we employ five widely used machine learning classifiers to construct blood cell classification models. Experimental results on two public blood cell datasets and a private real dataset demonstrate that our proposed DoRL achieves a new state-of-the-art cross-domain performance, surpassing existing methods by a significant margin. The source code can be available at the URL 
 (\url{https://github.com/AnoK3111/DoRL}).
\end{abstract}


\section{Introduction}
\begin{figure}[tb]
    \centering
    \includegraphics[width=1.0\linewidth]{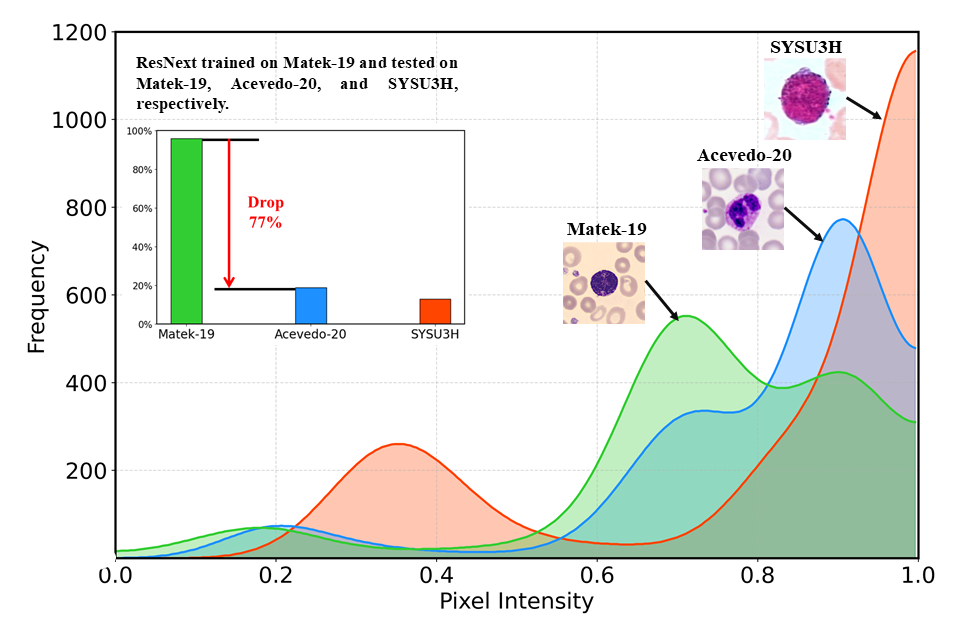}
    \caption{The grayscale distribution of different blood cell image datasets. The blood cell image datasets often exhibit domain shift issues, resulting in the deterioration of model generalization performance. For instance, when trained on the Matek-19 dataset, ResNext can achieve impressive performance on the source dataset. However, its performance deteriorates significantly when applied to unseen datasets (Acevedo-20 and SYSU3H).}
    \label{issue}
\end{figure} 

\IEEEPARstart{B}{lood} cell analysis is indispensable in medical diagnostics, serving as a fundamental method for diagnosing a wide range of diseases. It provides vital assistance to clinicians in identifying and monitoring conditions including leukemia, anemia, infections, autoimmune disorders, and various hematological diseases \cite{anilkumar2020survey}. Accurate and timely classification of blood cells is pivotal in clinical practice, enabling early detection and intervention for blood cell disorders\cite{wang2021research}. This proactive approach improves patient health conditions through timely recognition and treatment of blood-related ailments. Therefore, blood cell analysis stands at the forefront of medical practices, safeguarding patient health through proactive disease management strategies.

In clinical practice, traditional blood cell identification is performed manually by experienced physicians and specialists using a microscope. This process is repetitive, time-consuming, and labor-intensive, leading to potential delays in diagnosis and increased workload for medical professionals. Recently, significant advancements have been achieved in deep learning techniques for solving real-world classification tasks in the field of computer vision \cite{liu2023revisiting} \cite{van2023pdisconet}. Consequently, some researchers attempted to apply machine learning techniques to blood cell classification \cite{xie2017aggregated}\cite{acevedo2021new}, achieving promising results on specific datasets. However, blood cell image datasets often exhibit the issue of domain shift, which is caused by various factors, such as differences in illumination, microscope settings, camera resolution, and staining protocols (see Fig. \ref{issue}). These variations can significantly degrade model performance, rendering established methods ineffective and necessitating re-annotation and retraining.

Regarding the issue of domain shift, unsupervised domain adaption servers as an effective solution by aligning different domains in an unsupervised manner. It can be implemented through diverse approaches, encompassing moment matching and adversarial training. Moment matching is a technique that aims to minimize the discrepancy in feature distributions between source and target domains, including DeepCoral \cite{sun2016deep} and SWD \cite{lee2019sliced}. Adversarial training, following the architecture of Generative Adversarial Networks (GANs) \cite{creswell2018generative}, is designed to extract the domain-invariant features through adversarial processes, like DANN \cite{ganin2016domain} and CyCADA \cite{hoffman2018cycada}. In addition to the above methods, researchers have proposed other approaches for unsupervised domain adaption recently, such as MCC \cite{jin2020minimum} and LEAD \cite{sanqing2024LEAD}. However, most of these approaches overlook the artifacts in the images that may deteriorate the models' generalization performance.

\begin{figure*}[tb]
    \centering
    \includegraphics[width=1.0\linewidth]{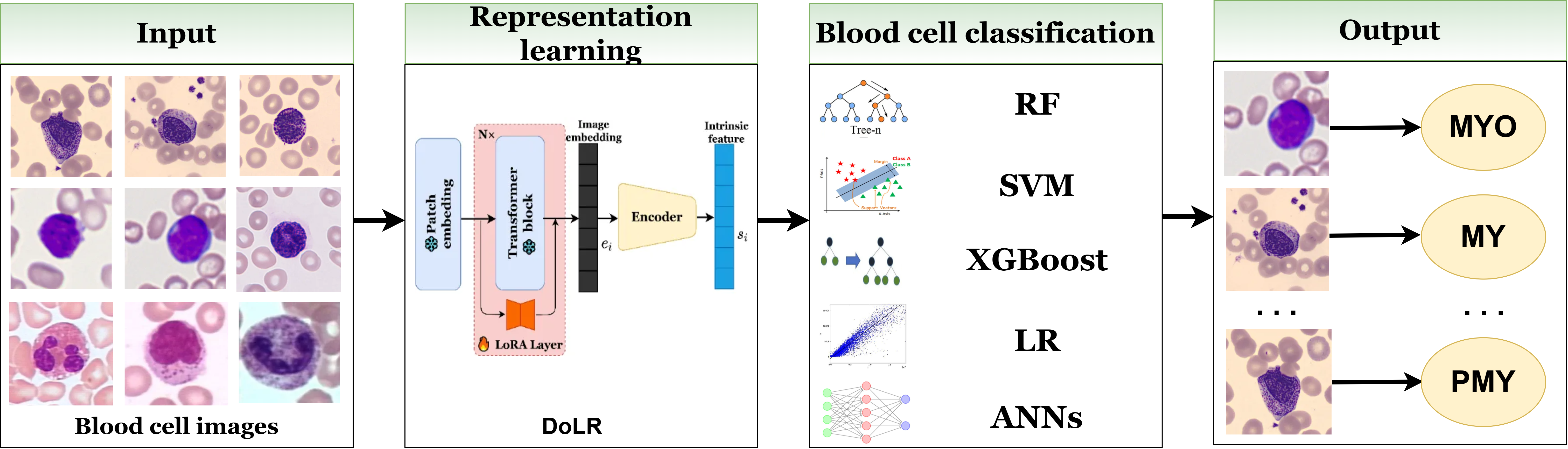}
    \caption{The pipeline of our proposed blood cell classification method. It comprises two primary components: domain-invariant representation learning and blood cell classification. In the stage of domain-invariant representation learning, DoRL is employed to extract domain-invariant features from blood cell images. To evaluate the effectiveness of these domain-invariant features, five widely used machine learning classifiers are employed to construct blood cell classification models.}
    \label{pipe}
\end{figure*} 

Recently, significant advancements have been made in large-scale foundation models for natural language processing tasks, such as LLaMA \cite{touvron2023llama} and GPT-4 \cite{nori2023capabilities}, as well as in computer vision tasks, including SAM \cite{kirillov2023segment} and SegGPT \cite{wang2023seggpt}. Researchers have also ventured into developing large-scale foundation models specific to the medical domain, such as Med-SAM\cite{ma2023segment} and SAMed \cite{zhang2023customized}. Drawing inspiration from the remarkable versatility of foundation models, we propose a novel framework of domain-invariant representation learning via Segment Anything Model (SAM) for blood cell classification, referred as DoRL. The DoRL framework encompasses two main components: a LoRA-based Segment Anything Model (LoRA-SAM) and a cross-domain autoencoder (CAE), enabling the model to extract domain-invariant representations from different domains unsupervisedly. Specifically, we first employ the SAM to learn image embeddings and segment blood cell images with the aid of low-rank adaption (LoRA) \cite{hu2021lora} for finetuning. Additionally, we introduce the CAE to extract domain-invariant representations from image embeddings while eliminating artifacts in blood cell images. Finally, we construct blood cell classification models leveraging five widely used machine learning algorithms: Random Forest (RF), Support Vector Machine (SVM) with linear and poly kernels, Logistic Regression (LR), Artificial Neural Network (ANN), and Extreme Gradient Boosting (XGBoost). Experimental results on two public datasets, as well as a private real dataset obtained from the Third Affiliated Hospital of Sun Yat-sen University named SYSU3H (IRB No. RG2023-265-01), demonstrate that our proposed DoRL outperforms existing methods, achieving state-of-the-art results by a significant margin. In summary, the primary contributions of our work can be summarized as follows:

\begin{itemize}
    \item We introduce the large-scale foundation model SAM into the blood cell identification task to learn general image embeddings and segment the blood cell images, with LoRA for fine-tuning.
    
    \item We propose a novel framework combining LoRA-SAM and CAE to learn rich domain-invariant features from blood cell images while effectively muting the domain-specific artifact of images.  
    
    \item Experiment results conducted on two public blood cell datasets and a private blood cell dataset show that our proposed DoRL can achieve a new cross-domain state-of-the-art result in blood cell classification.
\end{itemize}

The remainder of this paper is structured as follows. Section \ref{rw} reviews the related work on blood cell classification, large-scale foundation model, and unsupervised domain adaption. Section \ref{method} presents detailed descriptions of the proposed DoRL. The experimental settings and results are discussed in Section \ref{exp}, followed by the conclusion of this paper in Section \ref{con}.

\section{Related Work}
\label{rw}
\subsection{Blood Cell Classification}
Recently, blood cell classification as a special image classification task has attracted much attention from researchers. The advent of deep learning has led to breakthroughs in various domains like natural language processing \cite{xin2020deebert} \cite{yang2024seed}, computer vision \cite{munir2024greedyvig} \cite{yi2024leveraging}, and medical image analysis \cite{xie2024pairaug} \cite{kalantidis2024label}. Traditional deep learning methods such as deep neural networks \cite{matek2021highly}, convolutional neural networks \cite{acevedo2021new}, and ResNext \cite{xie2017aggregated} have been applied to blood cell classification tasks. Despite their initial success, these methods often experience a rapid decline in performance when applied to unseen datasets due to domain shifts. These domain shifts arise from various external factors, including differences in microscope settings, staining solutions, and lighting conditions. To address these challenges, Salehi et al. \cite{salehi2022unsupervised} proposed an adapted autoencoder based on an R-CNN architecture to extract features across three different datasets, aiming to bridge the gap between diverse domains. While their approach demonstrates improved performance in multi-domain blood cell classification, significant room remains to further enhance model's the generalizability and robustness.

\subsection{Large-scale Foundation Model}

The remarkable extension ability of the Transformer makes it possible for researchers to construct large-scale foundation models with billions of parameters. Large-scale foundation models first bring breakthroughs in the field of natural language processing, such as LLaMA \cite{touvron2023llama}, GPT-4 \cite{nori2023capabilities}, and FBI-LLM\cite{ma2024fbi}. They all demonstrate promising performance and applicability in natural language processing tasks compared to previous methods. Recently, large-scale foundation models have been widely applied to computer vision tasks, including SAM \cite{kirillov2023segment}, IDA-VLM \cite{ji2024ida}, and EVF-SAM \cite{zhang2024evf}. These methods achieved significant success in general computer vision tasks and their astonishing generalization ability received a huge amount of attention. Additionally, some researchers have applied these models specific to the medical domain like Med-SAM \cite{ma2023segment} and SAMed \cite{zhang2023customized}. Drawing inspiration from the outstanding versatility of large-scale foundation models, we present LoRA-SAM which is designed with the assistance of LoRA \cite{hu2021lora} to customize SAM for blood cell image segmentation.

\subsection{Unsupervised Domain Adaption}

Unsupervised domain adaptation addresses the challenge posed by the different domain data distributions between train and test datasets. The objective of unsupervised domain adaptation is to develop machine learning models that can generalize from source domains to target domains, effectively minimizing the discrepancies between various domains. Exiting unsupervised domain adaptation methods can be broadly categorized into two paradigms, encompassing moment matching \cite{tranheden2021dacs}\cite{kang2019contrastive} and adversarial training \cite{ganin2016domain}\cite{hsu2020progressive}.

Moment matching is a technique aimed at minimizing the discrepancy in feature distributions between source and target domains. For instance, Deep Coral \cite{sun2016deep} aligned the second-order statistics of the source and target distributions via nonlinear transformations. Additionally, DNN \cite{long2015learning} leveraged maximum mean discrepancy (MMD) \cite{li2018domain} for mean embedding matching to learn statistically transferable features, while SWD \cite{lee2019sliced} introduced sliced wasserstein distance to capture inherent dissimilarities in classifier outputs. 

Adversarial training, inspired by GANs \cite{creswell2018generative}, aims at learning domain-invariant features through adversarial processes. For example, DANN \cite{ganin2016domain} introduced a domain discriminator to distinguish source domains and target domains so that the features extracted by the feature extractor can deceive the domain discriminator and ensure the classification task effect of the source domain. Motivated by Conditional GANs \cite{mirza2014conditional}, many methods have been proposed to extract the domain-invariant features based on this architecture, including CyCADA \cite{hoffman2018cycada} and DTA \cite{lee2019drop}. 

In addition to moment matching and adversarial training, there are still other approaches for unsupervised domain adaption, such as MCC \cite{jin2020minimum} and LEAD \cite{sanqing2024LEAD}. Regardless of their effectiveness in addressing the issue of domain shift, most of the existing methods overlook the artifacts in the images that may deteriorate the model's generalization capabilities, especially in the task of blood cell classification.

\begin{figure*}[tb]
    \centering
    \includegraphics[width=1.02\linewidth]{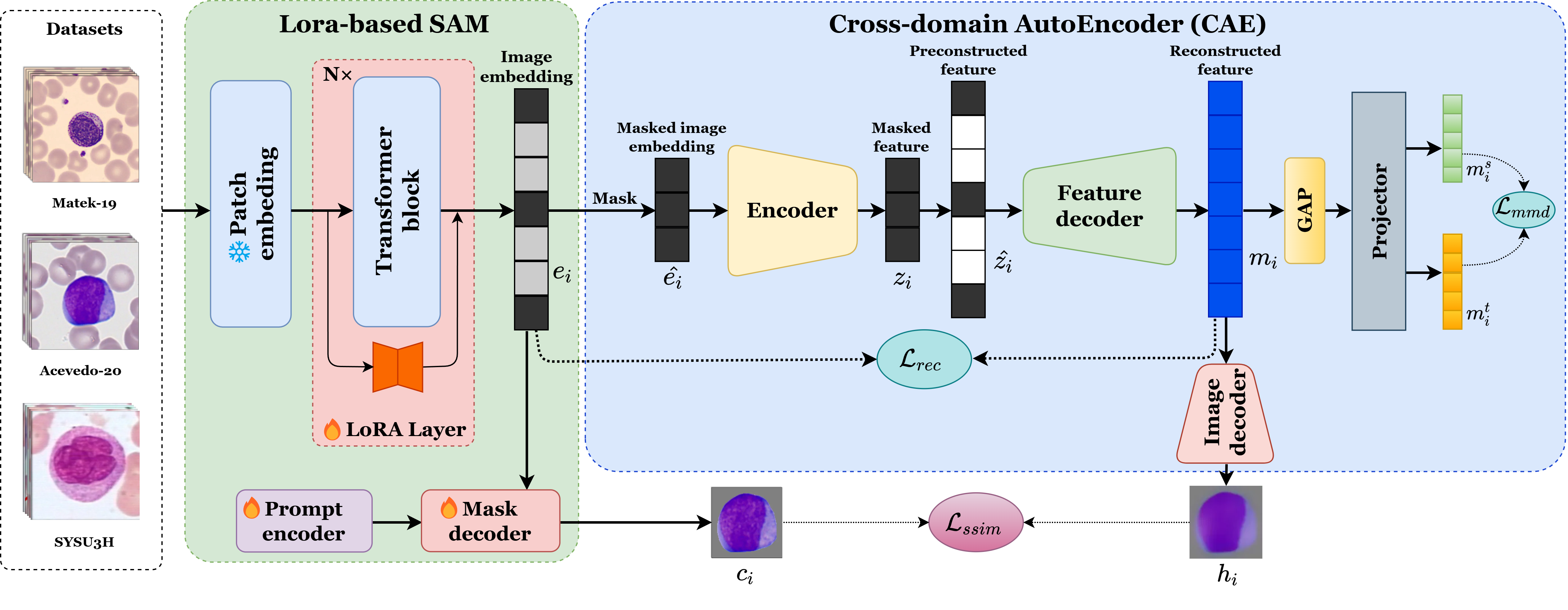}
    \caption{The architecture of our proposed DoRL. It consists of two key components: the LoRA-SAM and a CAE. The LoRA-SAM component is tasked with learning image embeddings from blood cell images and performing segmentation on these images. Subsequently, the segmented blood cell images and their corresponding embeddings are input into the CAE, which extracts domain-invariant features from various blood cell datasets while eliminating image artifacts.}
    \label{DoRL}
\end{figure*}

% Snowflake refers to the frozen parameters while flame refers to the learnable parameters.
\section{Methodology}
\label{method}
Fig. \ref{pipe} demonstrates the pipeline of our proposed DoRL, including two stages: domain-invariant representation learning and blood cell classification. In the stage of representation learning, we propose a novel architecture termed DoRL, consisting of two components: a LoRA-Based Segment Anything Model (\emph{i.e.}, LoRA-SAM) and a cross-domain autoencoder (\emph{i.e.}, CAE) to extract domain-invariant representations from blood cell images while suppressing domain-specific artifacts (see Section \ref{rl}). To evaluate the effectiveness of the extracted features, we leverage five widely used machine learning classifiers for blood cell classification, including RF, SVM, LR, ANNs, and XGBoost (see Section \ref{bcc}).

Formally, given a blood cell image \(x \in \mathbb{R}^{H \times W \times C}\) with spatial resolution \(H \times W\) and \(C\) channels, our objective is to predict the corresponding labels \(\hat{y}\) as accurately as possible, approximating the ground truth \(y\). Specifically, let \(D = \{D_1, D_2, ..., D_M\}\) (with \(M > 1\)) represent different blood cell datasets and the training dataset \(D_M\) can be defined as \(D = \{(x_i, y_i) \mid i = 1, 2, \ldots, N\}\), where N is the sample sizes and \(y_i \in \{1, 2, \ldots, L\}\) denotes the label of corresponding blood cell category, and \(L\) is the number of blood cell category. Our model denoted as \(f_{\text{DoRL}}(\cdot)\), is employed for domain-invariant feature extraction, yielding the domain-invariant features \(s_i\) from the blood cell images \(x_i\). Subsequently, utilizing these extracted features as input, we construct blood cell classification models \(f_{\text{BCCMs}}(\cdot)\), which generate the corresponding predicted labels \(\hat{y}\).

% Formally, given a blood cell image \(x \in \mathbb{R}^{H \times W \times C}\) whose spatial resolution is \( H \times W \) and the channel number is \(C\), our objective is to predict its corresponding labels \(\hat{y}\) as close to the ground truth \(y\) as possible. Specifically, let \(D = \{D_1, D_2, ..., D_M\}\)(with \(M > 1\)) represent different blood cell datasets and the training dataset \(D_M\) can be defined as \(D = \{(x_i, y_i) \mid i = 1, 2, \ldots, N\), where N is the sample sizes and \(y_i \in \{1, 2, \ldots, C\) denotes the label of corresponding blood cell category, and \(C\) is the number of blood cell category. Additionally, let \(f_{DoRL}= (\cdot)\) denote our model for domain-invariant feature extraction and the outputting corresponding to the blood cell domain-invariant features \(z_i\). Finally, using the extracted domain-invariant feature as input, we build blood cell classification models \(f_{BCCM}= (\cdot)\), which output the corresponding labels \(\hat{y}\).

\subsection{Damian-invariant Representation Learning}
\label{rl}
Fig. \ref{DoRL} shows the DoRL network architecture, which comprises two main components: LoRA-SAM and CAE. Initially, LoRA-SAM is employed to learn the image embeddings and segment the blood cell images. These embeddings are then fed into the CAE, aiming to extract the domain-invariant features and reconstruct the segmented blood cell images. Since image embeddings are blood cell-specific, this process enables the CAE to reconstruct only the blood cells, effectively eliminating artifacts that could degrade the quality of the domain-invariant features.

% In this section, we introduce a novel framework called DoRL, designed to extract domain-invariant features from blood cell images while mitigating domain-specific artifacts. As depicted in Fig. ?, the DoRL architecture comprises two primary components: LoRA-SAM and cross-domain autoencoder. Initially, we introduce LoRA-SAM to learn the image embedding and obtain segmented images of blood cell images. The cross-domain autoencoder utilizes extracted image embedding as input, aiming to reconstruct segmented blood cell images. By doing this, the cross-domain autoencoder can effectively extract domain-invariant features while discarding artifacts such as red blood cells.

\subsubsection{LoRA-based SAM}
In the central part of Fig. \ref{DoRL}, we introduce LoRA-SAM to learn the embeddings from blood cell images and generate the corresponding segmentation mask. Despite the impressive zero-shot generalization capabilities of SAM, its performance degrades significantly when directly applied to blood cell image segmentation tasks, even using various prompt schemes. Therefore, we finetune the SAM's parameters to enhance its effectiveness in segmenting blood cell images. Compared with finetuning all the parameters of SAM, we employ LoRA, a resource-efficient and precise finetuning approach that updates only a small fraction of the parameters during training. This strategy not only reduces the overhead computation but also achieves remarkable performance in blood cell segmentation. Specifically, we freeze the transformer blocks in SAM to keep the parameters $W$ fixed and introduce trainable LoRA Layers $\Delta W$. The structure of the LoRA layer comprises two linear layers, denoted as \(A \in {R^{r \times {C_{in}}}}\) and \(B \in {R^{{C_{out}} \times r}}\), where \(r \ll \min \{ {C_{in}},{C_{out}}\}\). Here, $r$ represents the domain-invariant dimension and is empirically set to 4. Consequently, the parameters of the update layers can be defined as follows:

\begin{equation}
    \hat{W} = W + \Delta W = W + BA,
\end{equation}

% Since the multi-head self-attention mechanism utilizes cosine similarity to determine the focus region, LoRA can be implemented in the projection layer of the query, key, or value to influence the attention score. Following the study of SAMed \cite{zhang2023customized}, we apply LoRA to the query and value projection layers, which can achieve better performance. Therefore the finetuning strategy of multi-head self-attention can be defined as follows:

% \begin{equation}
%     \text{Att}(Q, K, V) = \text{softmax}\left(\frac{QK^T}{\sqrt{C_{out}}} + B\right)V
    
% \end{equation}

The prompt encoder, along with the subsequent mask decoder, is composed of lightweight transformer layers, allowing for direct fine-tuning without the need for LoRA layers. Owing to SAM's remarkable few-shot adaptation ability, LoRA-SAM can attain outstanding segmentation performance by training on merely 1\% of blood cell images with masked annotations. Upon acquiring the blood cell segmentation masks from the mask decoder, we implement a post-processing step that crops the original image to preserve only the regions corresponding to the mask areas, while setting the grayscale values outside the masked regions to 128. Consequently, given an image $x_{i}$, we derive the image embedding \(e_{i}\) and segmented blood cell image \(c_{i}\) generated by LoRA-SAM, defined as followed,

\begin{equation}
    e_{i}, c_{i} = f_{LoRA-SAM}(x_{i};A, B),
\end{equation}

\subsubsection{CAE}
\label{cd-auto}

In the right part of Fig. \ref{DoRL}, we design a cross-domain autoencoder (\emph{i.e.}, CAE) to extract the domain-invariant representation \(s_i\) from image embedding \(e_{i}\) and eliminate artifacts in the image by reconstructing the segmented blood cell image \(c_{i}\). Specifically, the CAE consists of three modules: an encoder, a feature decoder, and an image decoder. Inspired by the study of masked autoencoders (MAE) \cite{he2022masked}, we first mask 75\% random patches of the image embedding \(e_{i}\) and reconstruct it from the latent representation and mask tokens, using the encoder and feature decoder modules. The encoder includes twelve vision transformer layers, while the lightweight feature decoder comprises only four vision transformer layers. This asymmetric encoder-decoder architecture enables efficient and effective model training. Additionally, we introduce an image decoder to reconstruct the segmented blood cell image \(c_{i}\), aiming to suppress the domain-specific information of the images. Therefore, the proposed CAE can be formulated as,

\begin{eqnarray}
    z_{i}&=&f_{enc}(\hat{e_{i}};\theta_{e}), \\
    m_{i}&=&f_{dec}^{feat}(\hat{z_{i}};\theta_{dt}), \\
    h_{i}&=&f_{dec}^{img}(m_{i};\theta_{di}),
\end{eqnarray}

\noindent
where $\hat{e_{i}}$ is the masked image embedding, $z_i$ is the masked feature generated by encoder, $\hat{z_i}$ is the masked feature with mask token for reconstruction, $m_{i}$ is the feature reconstructed by the feature decoder, and $h_i$ is the images reconstructed by the image decoder. Here, $\theta_{e}$, $\theta_{dt}$, and $\theta_{di}$ represent the parameters of the encoder, feature decoder, and image decoder, respectively. 

The encoder and feature decoder are designed to learn domain-invariant features by minimizing the discrepancy between different domains and reconstructing the feature embeddings. This is achieved by optimizing the following loss function:

\begin{eqnarray}
    \mathcal{L}_{\text{rmmd}}&=&\beta\mathcal{L}_{rec} + \mathcal{L}_{mmd},\\
    \mathcal{L}_{\text{rec}}&=&\frac{1}{n} \sum_{i=1}^{n} (e_i - m_i)^2,\\
    \mathcal{L}_{\text{mmd}}&=&\frac{1}{n_s} \sum_{i=1}^{n_s} ({m}_i^s) - \frac{1}{n_t} \sum_{j=1}^{n_t} ({m}_j^t),
\end{eqnarray}
    % \mathcal{L}_{rmmd}&=\frac{1}{n} \sum_{i=1}^{n} (e_i - m_i)^2 + \mathcal{L}_{mmd}
\noindent
where \(\mathcal{L}_{rec}\) is the feature reconstruction loss function, \(\beta\) is a trade-off hyperparameter between the two loss terms, and \(n\) is the size of the mini-batch of image embeddings. Additionally, we utilize the maximum mean discrepancy (MMD) \cite{li2018domain} loss function to minimize the discrepancy between different datasets. Here, \({m}_i^s\) and \({m}_i^t\) represent the features from the source and target domains, and \(n_s\) and \(n_t\) denote the number of features from the source and target domains, respectively. It is significant to note that we first utilize a global average pooling (GAP) layer and a linear projection layer the dimension of the feature \({m}_i\) before the optimizing \(\mathcal{L}_{\text{mmd}}\), which can effectively address the overfitting issue and accelerates the convergence of the DoRL model training.

To mute the domain-specific artifact of images and learn rich domain-invariant features, we utilize the Structure Similarity Index Measure (SSIM) \cite{1284395} to measure the similarity between the reconstructed images $h$ generated by feature decoder and post-processed images $c$ generated by LoRA-SAM. Formally, the similarity loss $\mathcal{L}_{ssim}$ is defined as:

\begin{equation}
\label{ssim}
\mathcal{L}_{ssim}=1 - \frac{{(2{\mu _h}{\mu _p} + {c_1})(2{\sigma _{hp}} + {c_2})}}{{(\mu _h^2 + \mu _p^2 + {c_1})(\sigma _h^2 + \sigma _p^2 + {c_2})}},
\end{equation}

\noindent
where $\mu$ and $\sigma$ are the mean and variance of the images and \(c_1\), \(c_2\) are small constants for numerical stability. Therefore, the total objective loss $\mathcal{L}$ of DoRL can be defined as,

\begin{equation}
\mathcal{L} =  \mathcal{L}_{ssim} +\lambda\mathcal{L}_{rmmd}.
\end{equation}

\noindent 
where $\lambda$ denotes a trade-off hyperparameter between the two loss terms. Finally, the domain-invariant feature $s_i$ can be obtained from the trained encoder $f_{enc}(\cdot)$.

\subsection{Blood Cell Classification}
\label{bcc}
In this section, using the learned domain-invariant features $s_i$ by DoRL as input, we utilize the five widely used machine learning methods of RF, SVM, LR, ANNs, and XGBoost to build blood cell classification models \(f_{\text{BCCMs}}(\cdot)\), which predict the corresponding blood cells category \(\hat{y}\). On the other hand, these classification models can further verify the cross-domain feature learning capability of our proposed DoRL.

\subsubsection{RF}
RF is a powerful machine learning algorithm for classification that builds an ensemble of decision trees. It utilizes bootstrap aggregation (bagging) and random feature selection to improve accuracy and mitigate overfitting. By averaging predictions from multiple trees, RF enhances model performance and generalization beyond that of individual decision trees.

\subsubsection{SVM}
SVM is a robust classification algorithm that identifies the optimal hyperplane to maximize class separation in high-dimensional spaces. It employs kernel functions (e.g., linear, polynomial, RBF) to handle non-linearly separable data by transforming it into higher dimensions. By minimizing a regularization term and hinge loss, SVM effectively balances margin maximization with error reduction, offering strong generalization and resilience to overfitting.

\subsubsection{LR}

LR is a fundamental binary classification algorithm that estimates class probabilities by applying the logistic function to a linear combination of features. It is trained to minimize prediction errors using methods like gradient descent and excels with linearly separable data and moderate-sized datasets. LR is valued for its simplicity, interpretability, and ability to provide probabilistic predictions and feature importance insights.

\subsubsection{ANN}

ANN is an advanced model inspired by the human brain, employing interconnected neurons and activation functions to transform input data. Trained via backpropagation and gradient descent, ANN minimize loss functions to capture complex, non-linear patterns in data. Their versatility and capacity for handling large datasets make ANNs highly effective for classification tasks and accurate predictions.

\subsubsection{XGBoost}

XGBoost is a powerful and efficient classification algorithm that builds robust models by combining weak learners, typically decision trees, through gradient boosting. It incorporates techniques such as regularization, parallel processing, and tree pruning to enhance accuracy and prevent overfitting. Known for its scalability and precision, XGBoost excels at handling large datasets and complex classification tasks.

\begin{table*}[tb]
\belowrulesep=0pt
\aboverulesep=0pt
\tabcolsep=0.4cm
\center
\caption{Comparison with state-of-the-art methods on two public blood cell datasets and a private real dataset. The best performance is in \textbf{bold} and the second best is indicated with \underline{underline}.}
\resizebox{\textwidth}{!}{
\begin{tabular}{c c c c c c c c}

\hline
Trained on& \multicolumn{3}{c}{Matek-19} & \multicolumn{3}{c}{Acevedo-20} & \multirow{2}{*}{\centering Average(\%)}\\ 
\cmidrule(lr){0-0}\cmidrule(lr){2-4}\cmidrule(lr){5-7}
Tested on& Matek-19(\%) & Acevedo-20(\%) & SYSU3H(\%) & Matek-19(\%) & Acevedo-20(\%) & SYSU3H(\%)\\ 
\hline
\specialrule{0em}{2pt}{0pt}
ResNext\cite{matek2019human} &\textbf{96.41$\pm$0.2} &19.63$\pm$3.5 &13.51$\pm$1.5 &9.32$\pm$2.9 & \textbf{93.25$\pm$0.4} & 0.81 $\pm$1.0 & 38.82\\
DANN\cite{ganin2016domain} &82.37$\pm$4.3 &29.15$\pm$2.7 &16.71$\pm$2.5 &34.71$\pm$2.8 &58.42$\pm$1.5 &13.99$\pm$1.9 & 39.23\\
MCC\cite{jin2020minimum} &78.56$\pm$0.8 &14.23$\pm$2.0 &17.38$\pm$7.6 &22.79$\pm$2.4 &59.00$\pm$5.8 &18.62$\pm$1.3 & 36.76\\
AE-CFE-RF\cite{salehi2022unsupervised} &83.70$\pm$0.5 &21.90$\pm$0.4 &15.23$\pm$0.7 &45.10$\pm$0.5 & 65.20$\pm$0.5 &11.63$\pm$0.9 & 40.46\\
LEAD\cite{sanqing2024LEAD} &69.23$\pm$3.2 &52.54$\pm$5.6 &26.08$\pm$3.9 &44.36$\pm$3.8 & 64.69$\pm$3.9 &16.31$\pm$2.9 & 45.54\\
\hline
\specialrule{0em}{2pt}{0pt}
BC-SAM-RF\cite{Li2024Towards} & 89.54$\pm$0.3 & 23.36$\pm$0.1 & 21.03$\pm$0.5 & 47.21$\pm$1.2 & 72.30$\pm$0.8 &20.84$\pm$0.8 & 45.71\\ 
BC-SAM-XGBoost\cite{Li2024Towards} & 91.94$\pm$0.5 & 34.10$\pm$1.2 &23.14$\pm$0.4 & 47.63$\pm$1.9 & 76.24$\pm$0.7 &\underline{26.71$\pm$1.5}  & 49.96\\ 
BC-SAM-SVM (poly)\cite{Li2024Towards} & 91.46$\pm$0.4 & 29.46$\pm$0.3 & 28.39$\pm$1.1 & 54.22$\pm$0.7 &75.44$\pm$0.8 &24.65$\pm$1.6 & 50.60\\ 
BC-SAM-SVM (linear)\cite{Li2024Towards} & 92.22$\pm$0.5 & 37.12$\pm$1.3 & 25.68$\pm$0.5 & 37.02$\pm$0.8 &77.24$\pm$0.7 &26.46$\pm$1.8 & 49.29\\
BC-SAM-LR\cite{Li2024Towards} & 92.04$\pm$0.4 & 35.08$\pm$1.2 & 25.98$\pm$0.9 & 35.40$\pm$1.1 & 77.33$\pm$0.6 & 16.13$\pm$0.6 & 46.83\\
BC-SAM-ANN\cite{Li2024Towards} & 92.79$\pm$0.3 & 39.59$\pm$0.7 & \underline{28.94$\pm$1.2} & 38.84$\pm$4.3 & 79.01$\pm$0.5 & 22.66$\pm$1.6 & 50.31\\
\hline
\specialrule{0em}{2pt}{0pt}
DoRL-RF & 91.77$\pm$0.1 & 22.73$\pm$0.3 & 22.66$\pm$1.0 & 31.48$\pm$1.5 & 84.65$\pm$0.4 &23.26$\pm$4.4 & 46.09\\ 
DoRL-XGBoost & 94.54$\pm$0.1 & 50.53$\pm$2.1 &25.98$\pm$0.9 & 38.96$\pm$0.8 & 86.82$\pm$0.4 &24.83$\pm$3.4  & 53.61\\ 
DoRL-SVM (poly) & 94.64$\pm$0.3 & 35.44$\pm$0.3 & 23.75$\pm$1.3 & 32.80$\pm$1.5 &87.10$\pm$0.5 &12.39$\pm$3.3 & 47.69\\ 
DoRL-SVM (linear) & 94.25$\pm$0.4 & \textbf{63.18$\pm$1.8} & 27.61$\pm$2.7 & \underline{55.42$\pm$2.8} &88.05$\pm$0.5 &\textbf{33.65$\pm$3.6} & \textbf{60.36}\\
DoRL-LR & 94.95$\pm$0.3 & 50.75$\pm$5.0 & 26.16$\pm$5.9 & \textbf{62.91$\pm$4.3} & 88.16$\pm$0.2 & 24.53$\pm$6.2 & 56.38\\
DoRL-ANN & \underline{95.34$\pm$0.2} & \underline{61.99$\pm$1.8} & \textbf{29.73$\pm$5.0} & 49.64$\pm$4.8 & \underline{88.89$\pm$0.5} & 12.69$\pm$4.7 & \underline{57.91}\\
\hline
\end{tabular}
}
\label{res}
\end{table*}

\section{Results and Discussion}
\label{exp}
\subsection{Experimental Settings}
\label{es}
\subsubsection{Datasets}

\begin{figure}[tb]
    \centering
    \includegraphics[width=1.0\linewidth]{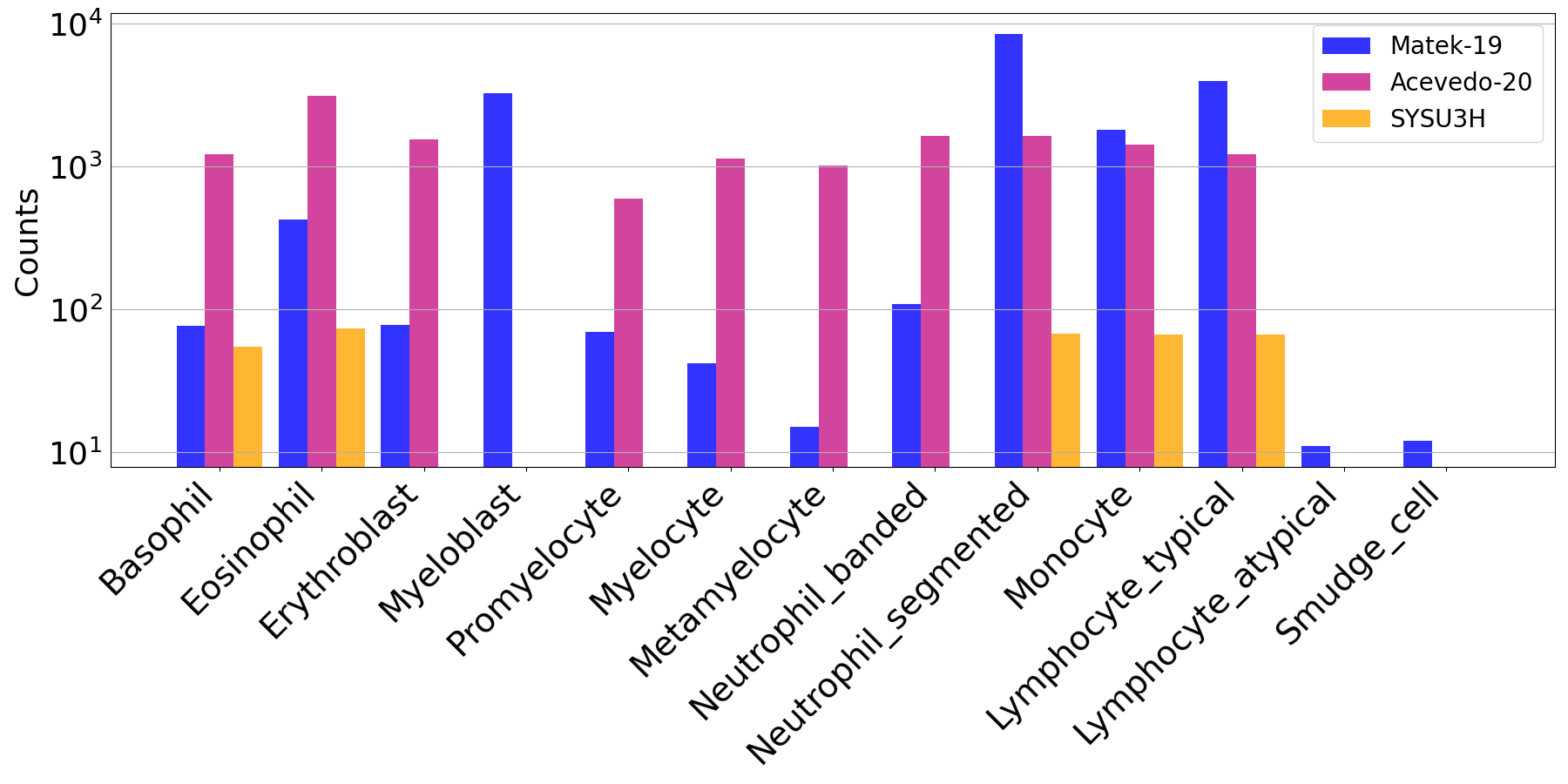}
    \caption{The statistic distribution of blood cell image datasets.}
    \label{data}
\end{figure} 

To evaluate the effectiveness of DoRL, we perform experiments on three blood cell image datasets: Acevedo-20 \cite{acevedo2020dataset}, Make-19 \cite{matek2019human}, and a private real dataset termed SYSU3H collected from The Third Affiliated Hospital of Sun Yat-sen University (IRB No. RG2023-265-01). These datasets were sourced from various laboratories and hospitals, ensuring the presence of domain shift among them. As shown in Fig. \ref{data}, we categorize the blood cell images into 13 classes based on their labels, and all of them are resized to \(224 \times 224\) pixels in our experiments.

\noindent
\textbf{Acevedo-20}: This dataset includes 17,092 blood cell images from healthy individuals, categorized into 10 classes. The images, with a resolution of \(360 \times 363\) pixels, were produced using a Sysmex SP1000i slide maker-stainer and stained with May Grünwald-Giemsa.

\noindent
\textbf{Matek-19}: This dataset contains over 18,000 annotated blood cell images across 13 distinct classes. Each image measures \(400 \times 400\) pixels, approximately \(29 \times 29\) micrometers. These samples were selected from a low-resolution pre-scan and scanned at 100x optical magnification with oil immersion using an M8 digital microscope.

\noindent
\textbf{SYSU3H}: This dataset is collected from The Third Affiliated Hospital of Sun Yat-sen University, with 331 blood cell images divided into 5 classes. All samples were stained with Wright-Giemsa and captured using a Leica ICC50 HD digital microscope camera.

\subsubsection{Evaluation Metrics}
\label{em}
The performance of the blood cell classifiers is evaluated using top-1 accuracy. Following the methodology outlined in \cite{salehi2022unsupervised}, we performed stratified train-test splits on the training datasets, reserving 20\% of the data as a hold-out test set. The remaining datasets are considered independent hold-out test domains for further evaluation. To evaluate the performance of our proposed DoRL, we utilize the 5-fold cross-validation for both single-domain and cross-domain datasets. For single-domain evaluation, we compute the average accuracy scores across the 5 folds to assess the framework's performance within a single domain. For cross-domain evaluation, the model is trained on one dataset and then validated through cross-domain testing. Specifically, each trained model is tested on two external datasets for every fold, and the cross-domain performance is determined by averaging the accuracy scores obtained on these external datasets across all folds.

\subsubsection{Implementation Details}
All experiments were performed on a computing server featuring an NVIDIA RTX6000 GPU with 24GB of video memory and dual Intel(R) Xeon(R) Gold 6248R CPUs with 256GB of system memory, running Ubuntu 20.04.5 LTS. The DoRL and blood cell classifiers were developed using Pytorch 1.9.1 and Scikit-learn 1.3.1. We employ the AdamW to optimize DoRL with an initial learning rate of 0.0005 and a weight decay rate of 0.05. In addition, warm-up and cosine learning schedules are presented to dynamically control the learning rate. LoRA-SAM is trained for 85 epochs while the CAE is trained for 10 epochs. As for the blood cell classifiers, we employ five machine learning models with their hyperparameters as follows: 1) RF: 200 estimators and 16 maximum depth; 2) SVM: linear and poly kernels; 3) LR: 500 maximum iterations; 4) ANNs: one hidden layer and 1000 maximum iterations with the number of hidden units set to 100; 5) XGBoost: default settings.

\subsection{Experimental Results}

The accuracy scores of various classification models on three blood cell datasets are listed in Table \ref{res}. Our proposed DoRL, incorporating SVM with a linear kernel, achieves the highest average classification accuracy of 60.36\%, significantly outperforming the state-of-the-art method, BC-SAM-SVM (poly) \cite{Li2024Towards}, by 9.76\%. Additionally, DoRL-ANN attains the second-highest average classification performance with an accuracy of 57.91\%, further demonstrating the effectiveness of DoRL in extracting domain-invariant features across different blood cell datasets.

When Matek-19 is used as the source dataset, DoRL-ANN shows competitive performance with an accuracy of 95.34\%, which is only 1.07\% lower than the top-performing model, ResNext \cite{matek2019human}. For cross-domain performance, DoRL-SVM (linear) and DoRL-ANN achieve the highest and second-highest accuracy scores on the Acevedo-20 dataset, with 63.18\% and 61.99\%, respectively. For the private real dataset SYSU3H, DoRL-ANN achieves the highest accuracy score of 29.73\%, surpassing the second-best method, BC-SAM-ANN, by 0.79\%.

When Acevedo-20 serves as the source dataset, DoRL-ANN performs comparably with an accuracy of 88.89\% on the source domain. In terms of cross-domain performance, DoRL-LR and DoRL-SVM (linear) achieve the highest and second-highest accuracy scores on the Matek-19 dataset, with 62.91\% and 55.42\%, respectively. For the private real dataset SYSU3H, DoRL-SVM (linear) achieves the highest accuracy score of 33.65\%, surpassing the second-highest method, BC-SAM-XGBoost, by 6.94\%.

Compared to existing baselines, ResNext \cite{matek2019human} achieves the highest accuracy on source domain datasets but exhibits significant performance degradation when tested on target domain datasets. Although current unsupervised domain adaptation models such as DANN \cite{ganin2016domain}, MCC \cite{jin2020minimum}, AE-CFE-RF \cite{salehi2022unsupervised}, and LEAD \cite{sanqing2024LEAD} demonstrate considerable improvements in target domain performance, most of these methods neglect images' artifacts that can undermine the models' generalization capabilities. In contrast, our proposed DoRL, based on the large-scale foundation model SAM, effectively extracts domain-invariant features from diverse datasets while mitigating the impact of images' artifacts. As illustrated in Table \ref{res}, DoRL, utilizing various machine learning methods, significantly outperforms previous unsupervised domain adaptation models, highlighting its superior capability in learning domain-invariant features from blood cell images.

\subsection{Visualization Analysis}

\begin{figure*}[tb]
    \centering
    \includegraphics[width=1.00\linewidth]{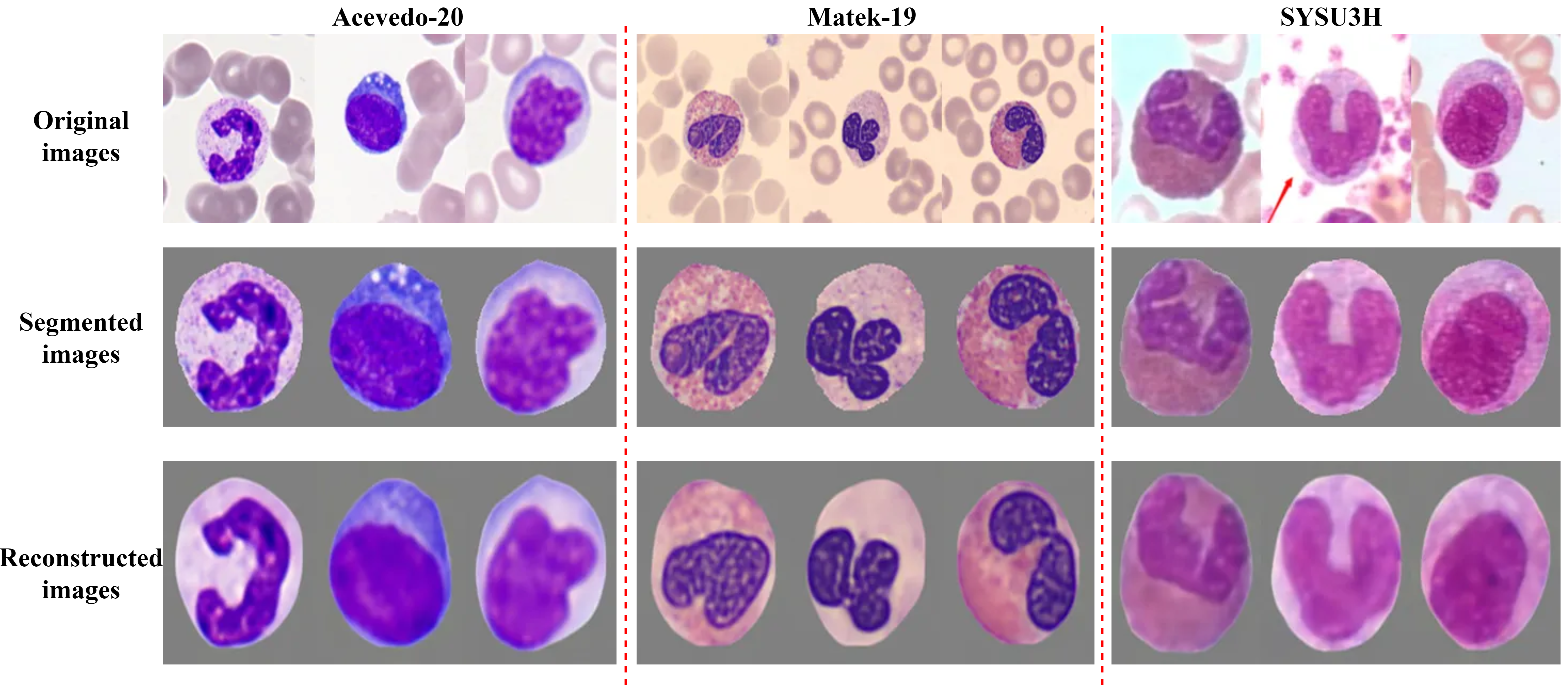}
    \caption{The visual presentation of original, segmented, and reconstructed images. Notably, the second row is the segmented images generated by LoRA-SAM and the third row is the images reconstructed by CAE. It is observed that our proposed DoRL can effectively eliminate artifacts and accurately reconstruct blood cells.}
    \label{fig:compared}
\end{figure*} 

\begin{figure*}[tb]
    \centering
    \includegraphics[width=1.00\linewidth]{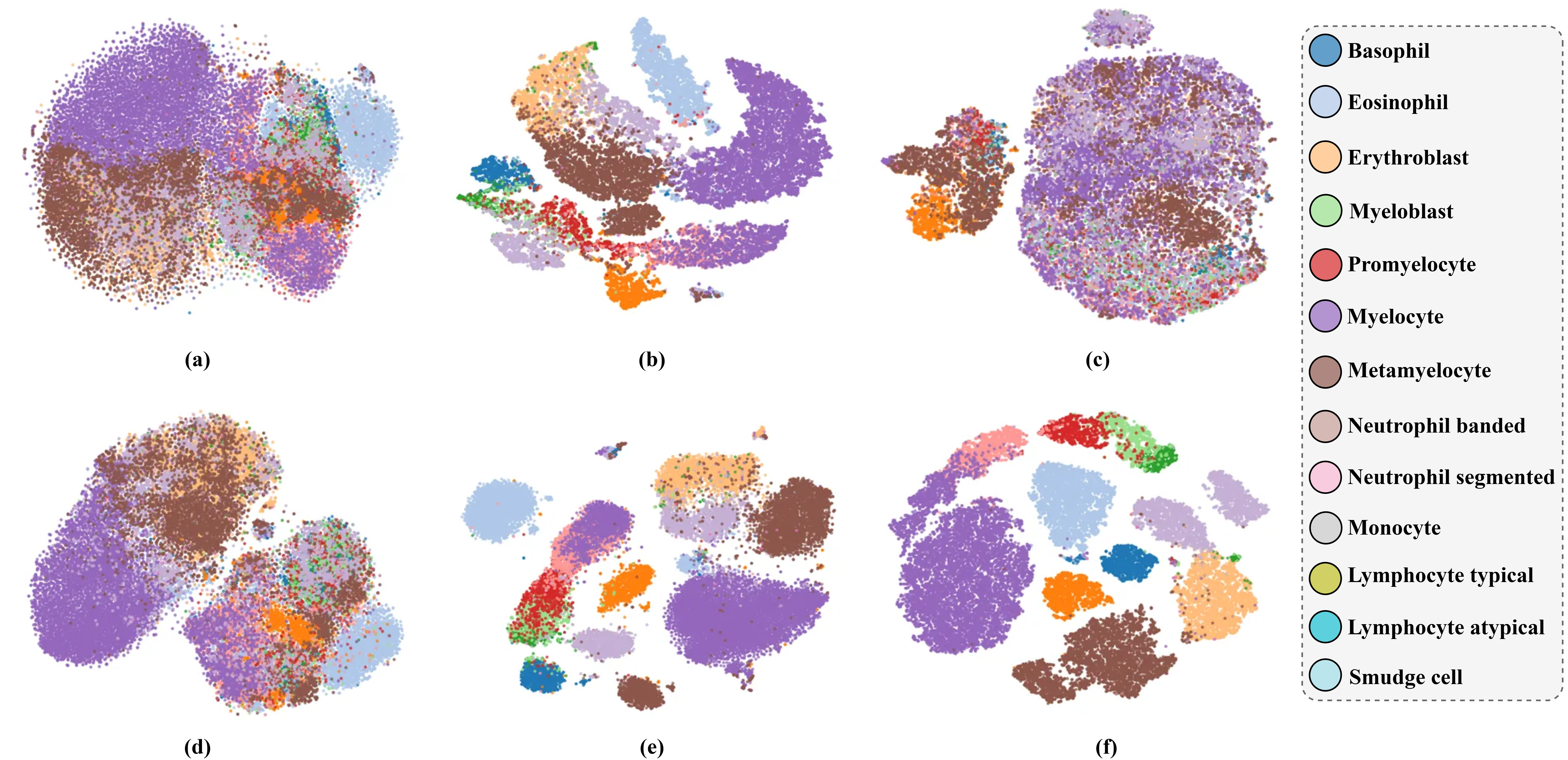}
    \caption{The t-SNE feature visualization. (a) ResNext. (b) DANN. (c) AE-CFE. (d) MCC. (e) LEAD. (f) our proposed DoRL. Compared to other models, our proposed DoRL effectively clusters data points with well-defined boundaries, demonstrating its robust representation learning capabilities.}
    \label{fig:tsne}
\end{figure*} 

The primary advantage of DoRL is its capacity to extract domain-invariant features while simultaneously eliminating artifacts. We evaluate the effectiveness of DoRL by comparing original, segmented, and reconstructed images across three blood cell datasets. Fig. \ref{fig:compared} presents a visual comparison of these images. The results demonstrate that LoRA-SAM achieves accurate blood cell image segmentation while CAE effectively reconstructs these segmented images, highlighting our proposed DoRL robustness in feature extraction and artifact removal.

Additionally, we leverage t-SNE \cite{van2008visualizing} to compute pairwise similarities in the latent space and visualize them in a lower-dimensional space, aligning the distributions using KL divergence. Fig. \ref{fig:tsne} demonstrates t-SNE visualizations for ResNext \cite{matek2019human}, DANN \cite{ganin2016domain}, AE-CFE \cite{salehi2022unsupervised}, MCC \cite{jin2020minimum}, LEAD\cite{sanqing2024LEAD}, and our proposed DoRL. It is worth noting that a model with better generalization should show more clustered points in the t-SNE plot. Compared to other models, DoRL demonstrates superior clustering of data points with clear boundaries, underscoring its enhanced generalization and robustness in multi-domain blood cell classification tasks.

\subsection{Hyperparameter Tuning}

\begin{figure*}[tb]
    \centering
    \includegraphics[width=1.00\linewidth]{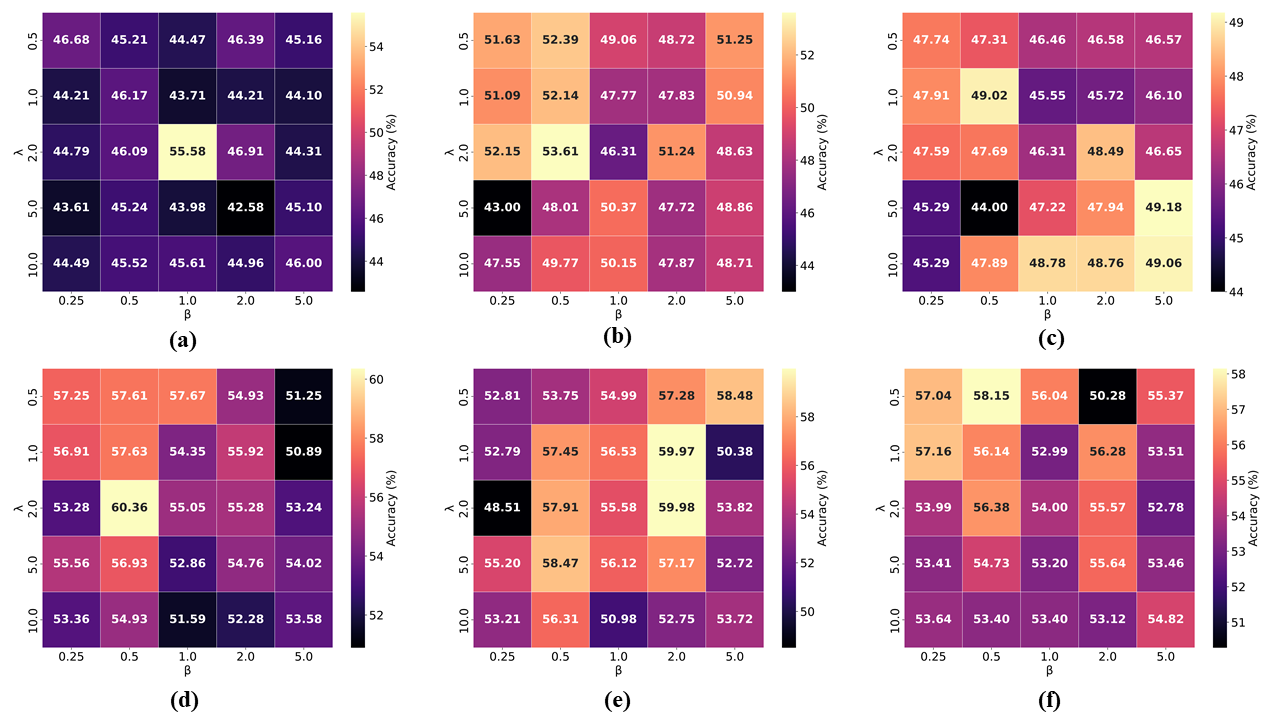}
    \caption{Hyperparameters tuning for DoRL. (a) RF. (b) XGBoost. (c) SVM (poly). (d) SVM (linear). (e) LR. (f) ANN. We select \(\beta = 0.5\) and \(\lambda = 2.0\) as the optimal parameters, which achieves the highest average accuracy score of 60.6\% with SVM (linear). }
    \label{fig:hyper}
\end{figure*} 

In this section, we systematically assess the performance of the DoRL method with various hyperparameter configurations to determine the optimal settings for our blood cell classification task. We experiment with different values for feature reconstruction loss weight (\(\beta\)) and the trade-off hyperparameter (\(\lambda\)) of \(\mathcal{L}_{rmmd}\), evaluating their performance on five widely used machine learning classifiers. This hyperparameter tuning is carried out under a rigorous framework outlined in Section \ref{es}. The results of hyperparameter tuning are the average classification performance of three different datasets, which is presented in Fig \ref{fig:hyper}.  We select \(\beta = 0.5\) and \(\lambda = 2.0\) as the optimal parameters, achieving the highest average accuracy score of 60.6\% with SVM (linear). Other classifiers also achieve competitive performance, with accuracy scores of 46.09\% using the RF classifier, 53.31\% using the XGBoost classifier, 47.69\% using the SVM (poly) classifier, and 56.38\% using the ANN classifier, respectively. These results highlight our selected parameters' effectiveness and our approach's robustness across various classifiers.

\subsection{Ablation Studies}

\begin{table*}[tb]
    \caption{Ablation experiment results of different variants models of DoRL conducted on three blood cell datasets.}   
    \resizebox{\textwidth}{!}{     
    \begin{tabular}{ccccccccccc}
    \hline    
     Variants & $\mathcal{L}_{rec}$ & $\mathcal{L}_{mmd}$ & $\mathcal{L}_{ssim}$ & RF & XGBoost & SVM (poly) & SVM (linear) & LR & ANN & Average\\
    \hline
    M1 &  &  &  & 33.94 & 37.48 & 35.57 & 45.07 & 45.20 & 42.51 & 39.96\\
    M2 & \checkmark &  &  & 44.20 & 47.90 & 47.39 & 53.56 & 53.56 & 54.56 & 50.19 \\
    M3 & \checkmark & \checkmark &  & 45.64 & 50.14 & 48.29 & 53.16 & 53.77 & 53.39 & 50.73 \\
    M4 & \checkmark &  & \checkmark & 46.27 & 52.42 & 46.88 & 53.73 & 55.05 & 56.22 & 51.76 \\
    M5 & \checkmark & \checkmark & \checkmark & 46.09 & 53.61 & 47.69 & 60.36 & 57.91 & 56.38 & 53.67 \\ 
      \hline
    \end{tabular}
    }
    \label{ab}
\end{table*}

To assess the contribution of different components within DoRL, we conduct extensive ablation studies using five widely used machine learning classifiers. The ablation experiment results for different variant models are the average scores on three public datasets and the experiments are conducted with the rigorous settings detailed in Section \ref{em}. The specific variants of our model are described as follows:
\begin{itemize}
    \item \textit{Variant M1}: Only using the image embedding obtained from LoRA-SAM for classification.
    \item \textit{Variant M2}: Variant M1 with the MAE \cite{he2022masked} architecture to enhance model's generalization capability. 
    \item \textit{Variant M3}: Variant M2 with MMD loss function $\mathcal{L}_{mmd}$ to reduce the discrepancy between different datasets.
    \item \textit{Variant M4}: Variant M2 with similarity loss function $\mathcal{L}_{ssim}$. 
    \item \textit{Variant M5}: Variant M2 with MMD loss function, \emph{i.e.,} our proposed DoRL.
\end{itemize}
Table \ref{ab} presents the ablation experiment results on three blood cell datasets using five classical machine learning classifiers. Compared with the DoRL without the component of CAE (\emph{i.e.,} variant M1), variant M2 with MAE architecture can enhance the average accuracy score by 10.23\%. After introducing the CAE module to learn discriminative feature representations, the performance of variant M2 can reach up to 44.20\% with the RF classifier, 47.90\% with the XGBoost classifier, 47.39\% with the SVM (poly) classifier, 53.53\% with SVM (linear) classifier, 53.26\% with LR classifier, 54.56\% with the ANN classifier, and the average score of 50.19\%. When incorporating the MMD loss function  $\mathcal{L}_{mmd}$ in variant M3, the average accuracy score increases by 0.54\%, demonstrating the benefit of narrowing the gap between different datasets. On the other hand, variant M4, which uses the similarity loss function $\mathcal{L}_{ssim}$ to mitigate image artifacts, achieves an average accuracy score of 51.76\%. This represents a 1.57\% improvement over variant M2. Notably, the DoRL (\emph{i.e.,} variant M5) with both MMD and similarity loss functions greatly facilitates the performance. It achieves an average accuracy score of 53.67\%, surpassing variants M3 and M4 by 3.48\% and 1.91\%, respectively. These ablation results underscore the synergistic benefits of the proposed components and highlight their superiority in multi-domain blood cell classification tasks.

\section{Conclusion}
\label{con}

In this study, we propose a novel blood cell classification framework termed DoRL, which comprises two main components: LoRA-SAM and CAE. Specifically, we first utilize SAM with the finetuning strategy of LoRA to learn image embeddings and obtain the segmented blood cell images. Additionally, we design CAE to extract the domain-invariant features leveraging the MMD loss function, while suppressing the artifact of the images by reconstructing the post-processed images. To evaluate the effectiveness of domain-invariant features, we leverage five widely used machine learning methods as classifiers for multi-domain blood cell classification. Experimental results on two public blood cell datasets and a private real-world dataset demonstrate that our proposed DoRL framework achieves a new state-of-the-art cross-domain performance, surpassing existing methods by a significant margin. The DoRL framework, with its novel integration of LoRA-SAM and CAE, provides a robust and effective solution for enhancing the generalization capabilities of blood cell classification models, potentially improving patient care and outcomes.

\section*{Acknowledgement}

This paper is an extension of the work presented at the IEEE-ISBI conference \cite{Li2024Towards}. We have further expanded the model architecture and supplemented the results with additional experiments and visualizations. Specifically, we introduce a random mask technique into our proposed DoRL to enhance the extraction of domain-invariant features from blood cell images. Additionally, we transition our encoder architecture from convolutional neural networks to vision transformer blocks and integrate a feature decoder to reconstruct the masked image embeddings, thereby improving our model's performance and robustness. Furthermore, we supplement the experimental analysis with a private real dataset (SYSU3H) and conduct comprehensive evaluations to validate the scalability and generalization capabilities of our model. We extend our gratitude to the organizers and reviewers of IEEE-ISBI for their invaluable feedback and support, which significantly contributed to the development of this work.

\bibliographystyle{IEEEtran}
% \bibliography{ref}

\end{document}